\documentclass[letterpaper, 10 pt, conference]{ieeeconf}  

\IEEEoverridecommandlockouts                              

\usepackage{graphicx} 
\usepackage{amsmath} 
\usepackage{amsfonts} 
\usepackage{amssymb}  
\usepackage{multirow}
\usepackage{xcolor}
 
\newcommand{\des}{\text{des}} 
\usepackage{hyperref}

\title{\LARGE \bf
A Comparison of Action Spaces for Learning Manipulation Tasks
}

\author{Patrick Varin, Lev Grossman, and Scott Kuindersma
\thanks{This work was funded by Schlumberger-Doll Research.} 
\thanks{School of Engineering and Applied Sciences, Harvard University,
        Cambridge, MA 02138, USA
        {\tt\small \{varin@g, lgrossman@college, scottk@seas\}.harvard.edu}}%
}

\begin{document}

\maketitle
\thispagestyle{empty}
\pagestyle{empty}

\begin{abstract}
Designing reinforcement learning (RL) problems that can produce delicate and precise manipulation policies requires careful choice of the reward function, state, and action spaces. Much prior work on applying RL to manipulation tasks has defined the action space in terms of direct joint torques or reference positions for a joint-space proportional derivative (PD) controller. In practice, it is often possible to add additional structure by taking advantage of model-based controllers that support both accurate positioning and control of the dynamic response of the manipulator. In this paper, we evaluate how the choice of action space for dynamic manipulation tasks affects the sample complexity as well as the final quality of learned policies. We compare learning performance across three tasks (peg insertion, hammering, and pushing), four action spaces (torque, joint PD, inverse dynamics, and impedance control), and using two modern reinforcement learning algorithms (Proximal Policy Optimization and Soft Actor-Critic). Our results lend support to the hypothesis that learning references for a task-space impedance controller significantly reduces the number of samples needed to achieve good performance across all tasks and algorithms.  
\end{abstract}

\section{Introduction}
Recent work in model-free reinforcement learning (RL) has demonstrated the ability to solve difficult high-dimensional problems in robotics \cite{OpenAI18, Levine16}. However, formulating a successful learning problem requires both expert knowledge and extensive experimentation to design the reward function, state space, and action space of the underlying Markov decision process (MDP). In many cases, we can take advantage of existing partial models of the problem (e.g., the equations of motion for a robot arm) to avoid learning dynamics that are well understood, and thus simplify learning. This paper is aimed at evaluating how learning performance can be improved across a set of example dynamic manipulation tasks by choosing action spaces that take advantage of model-based controllers. 

Two popular choices of action spaces in the robotics RL literature are direct torque control \cite{Lillicrap15, Watter15, Gu16, Levine15, Popov17, Chebotar17,Franceschetti18} and joint-space proportional derivative (PD) control \cite{OpenAI18, Peng18}. These action spaces are favorable because they are easy to implement, require relatively little information about the underlying system dynamics, and inject very little bias into the learning problem. On the other hand, these action spaces require learning to compensate for dynamic effects (e.g., inertial, gravitational, and centrifugal forces) that we are frequently able to model accurately. We expect that this may reduce learning performance, even for very simple manipulation tasks. Indeed, it is often necessary in practice to implement gravity compensation to learn successful manipulation strategies using torque control \cite{Kalakrishnan11a, Gu16}.

Johannink et al. have shown that formulating the learning task to learn residuals to a hand-crafted model-based controllers can improve learning efficiency \cite{Johannink18}. Another approach is to design the action space in terms of references for an underlying model-based controller. Inverse dynamics and task-space impedance control strategies have been successful for manipulation tasks because of their ability to achieve compliant interaction with the world, especially in the presence of uncertainty. Early success in high-tolerance peg insertion tasks, for example, was a result of the intelligent incorporation of mechanical compliance \cite{Whitney82}. In principle, RL algorithms that control joint torques directly can learn compliant manipulation strategies; however, it is not clear how this impacts sample efficiency and the quality of the resulting policy.



We hypothesize that taking advantage of model-based controllers and defining action spaces as reference inputs to these controllers can 1) improve the sample efficiency of learning manipulation tasks and 2) result in higher quality learned policies.  
In this paper, we provide preliminary evidence in support of this hypothesis using two modern on-policy (Proximal Policy Optimization, PPO \cite{Schulman17}) and off-policy (Soft Actor-Critic, SAC \cite{Haarnoja18, Haarnoja18a}) RL algorithms across four choices of action spaces for three simulated manipulation tasks (nail hammering, object pushing, and peg insertion).

\section{Background}
Early work in reinforcement learning for robotic manipulation was dedicated to learning compliant task-space control strategies. Gullapalli et al. learned a velocity controller that used position and force feedback to complete a high tolerance peg insertion task \cite{Gullapalli92}. Kim et al. used linear policies to learn the impedance gains around precomputed trajectories to complete a number of manipulation tasks \cite{Kim10}. Some related work learns full state trajectories, feedforward torques, and joint impedance for feedback control using PI$^2$ \cite{Kalakrishnan11a, Buchli11}.

Recent work in RL for manipulation has tended to take a more tabula rasa approach, focusing on learning policies that output joint torques directly or that output position (and velocity) references to an underlying PD controller. Direct torque control has been used to learn many physical and simulated tasks, including peg insertion, placing a coat hanger, hammering, screwing a bottle cap \cite{Levine15}, door opening, pick and place tasks \cite{Gu16}, and Lego stacking tasks \cite{Haarnoja18b}. Learning position and/or velocity references to a fixed PD joint controller has been used for tasks such as door opening, hammering, object placement \cite{Rajeswaran17}, Lego stacking \cite{Popov17}, and in-hand manipulation \cite{OpenAI18}.

The choice of feedback gains for joint-space PD controllers has a significant impact on how well a learned policy can perform---inappropriately chosen gains can result in poor tracking, instabilities, and chattering in the presence of time delays on physical systems, or even stiff system dynamics that require very small time steps in simulation. In response to this, Tan et al. \cite{Tan11} developed an implicit PD control scheme that can allow arbitrarily large gains. While this is a popular choice in the graphics community \cite{Geijtenbeek12, Liu13, Tan11a} and for learning in simulation \cite{Peng18, Andrews13, Liu16, Peng17a, Liu17}, the control law is non-causal and does not extend to physical systems.


While there is some work that formulates the action space in terms of a low-level Cartesian position controller for tasks such as block stacking \cite{Johannink18}, pushing, and pick and place \cite{Silver18}, there is surprisingly little work in reinforcement learning that uses impedance control to combine the ideas of task space control with the mechanical compliance necessary to perform delicate manipulation tasks under uncertainty, and no work to our knowledge that attempts a systematic comparison between alternative policy structures in this domain. 

The work presented here is related to work by Peng et al.~\cite{Peng17} that compares action spaces for robotic locomotion tasks in simulation. This paper compares torque, PD control, and a biologically inspired muscle activation model as action spaces for locomotion across a range of robot morphologies. This work showed that PD control is a better action space than direct torque control for locomotion. The focus on locomotion, however, precludes the use of controllers used in manipulation that take advantage of invertible dynamics (such as end-effector impedance controllers), which may offer significant performance benefits when learning interactive manipulation tasks. 
\section{Learning Algorithms}
Reinforcement learning problems are posed in the framework of Markov decision processes (MDP), which are defined by a set of states, $\mathcal{S}$, actions, $\mathcal{A}$, stochastic dynamics, $p(s_{t+1}|s_t, a_t)$, a reward function $r(s,a)$, and a discount factor, $\gamma$. The reinforcement learning objective is to compute the policy, $\pi^*(s,a)$, that maximizes the expected discounted sum of rewards, $\mathbb{E}_{s,a}\left(\sum_t \gamma^t r_t \right)$. Since we are interested in the choice of action space and its effect on learning performance, we vary that element of the MDP while keeping the states, rewards, and discount factor fixed.

We train all learning tasks with two state-of-the-art on- and off-policy reinforcement learning algorithms: Proximal Policy Optimization (PPO) and Soft Actor-Critic (SAC).  Implementations of both are made available by the \texttt{stable-baselines} \cite{stable-baselines} project, a software fork of the OpenAI \texttt{baselines} package.

\subsection{Proximal Policy Optimization (PPO)}
The Proximal Policy Optimization algorithm \cite{Schulman17} is an on-policy policy gradient \cite{Sutton00} method that uses an actor-critic framework to jointly learn the optimal policy as well as the optimal value function. Similar to Trust Region Policy Optimization (TRPO) \cite{Schulman15a}, PPO stabilizes policy training by penalizing large policy updates, an idea similar to trust-region methods or regularization techniques from the optimization literature. Because PPO is on-policy, each update must be computed with samples taken from the current policy, which can generally result in high sample complexity. 



\subsection{Soft Actor-Critic (SAC)}
The Soft Actor-Critic algorithm \cite{Haarnoja18, Haarnoja18a} is an off-policy reinforcement learning method that is based on soft Q-learning (SQL) \cite{Haarnoja17}. Unlike many RL algorithms, SAC optimizes a ``maximum entropy'' objective,
\begin{align}
     \mathbb{E}_{(s_t, a_t) \sim \pi}\left[\sum_{t} \gamma^t r(s_t,a_t) + \alpha \mathcal{H}\left(\pi(\cdot | s_t)\right) \right],
\end{align}
which encourages exploration according to a temperature parameter $\alpha$.

In this maximum entropy framework, the optimal policy is given by the soft Bellman equation which provides the basis for the SQL algorithm. SAC makes a number of improvements on SQL by automatically tuning the temperature parameter, $\alpha$, using double Q-learning, similar to the Twin Delayed DDPG (TD3) algorithm \cite{Fujimoto18}, to correct for overestimation in the Q-function, and learning not only the Q-functions and the policy but also the value function. Furthermore, because SAC is an off-policy algorithm it uses a replay buffer to reuse information from recent rollouts for sample-efficient training.


\section{Action Spaces}
There are a number of common control techniques employed by the RL community as well as in traditional manipulation. We analyze the performance of four common controllers on their performance in learning: direct torque control, proportional derivative (PD) control, an inverse dynamics (ID) controller, and an impedance controller. In traditional robotic control, the choice of controller to implement often considers a tradeoff between ease of implementation (torque and PD control are relatively simple), with the performance that can be gained by considering the dynamics of the system being controlled (ID and impedance control both attempt to compensate for the system dynamics). We hypothesize that we will see a similar tradeoff when designing the action space for an RL problem in terms of these controllers.

\subsection{Direct Torque Input}
The most common action space for reinforcement learning for robots in simulation maps actions directly to joint torques. This control strategy is trivial to implement and introduces minimal bias into the learning process in the sense that the policy can arbitrarily shape the robot's behavior within physical constraints. One criticism of controlling torques directly is that doing so requires the learned controller to compensate for the full dynamics of the robot, including gravity, Coriolis, and centrifugal forces. Additionally, it is possible for policies to output high-frequency torque signals, so care usually has to be taken to encourage smooth policy outputs before learned policies can be deployed on hardware. 

In our experiments, we found that even learning to compensate for gravity was difficult and required long training times when it did succeed. As a result, we augment the direct torque controller with a gravity compensation controller to improving training. This is a common technique \cite{Kalakrishnan11a, Gu16} and is easily justified by the fact that many robot arms have built-in gravity compensation controllers that must be treated as part of the closed-loop dynamics. Furthermore, because motors at proximal joints move more mass than those at distal joints, they tend to exert more torque. We found it beneficial for training to scale the torques at each joint by the cumulative mass of all of the child links. The resulting control law is
\begin{align}
    u = m_s \odot \pi(s) + g(q),
\end{align}
where $u$ are the control torques, $\pi$ is the policy, $s$ is the state, $m_s$ is a vector representing the subtree mass of each joint, and $g(q)$ is the position dependant gravity compensation control term. The operator $\odot$ is used to indicate element-wise vector multiplication.

\subsection{PD Control}
Another controller that makes minimal assumptions about the system dynamics is proportional derivative (PD) control. PD control can provide good tracking performance but at the cost of large gains, resulting in very stiff movements.

Similar to torque control, PD control suffers from scaling problems because the effective mass at each of the joints in an articulated body may span many orders of magnitude. This requires very different gains across each of the joints. In our experiments, we scale the proportional gains, $K_p$, by the subtree mass, and we choose the derivative gains, $K_d$, to be either the approximate critical damping gains or the maximum stable damping gains
\begin{align}
    K_p &= m_s \odot \tilde K_p\\
    K_d &= \min \left(\Delta t K_p, 2 \sqrt{m_s\odot K_p} \right),
\end{align}
where $\tilde K_p$ is a vector of unscaled proportional gains and $\Delta t$ is the simulation timestep.

The resulting control law is
\begin{align*}
    u = K_p(q_\des - q) + K_d(\dot q_\des - \dot q),
\end{align*}
where $q_\des$ and $\dot q_\des$ make up the action space for the controller.

\subsection{Inverse Dynamics Control}
The dynamics of an articulated body system can be written in terms of the manipulator equation,
\begin{align}
    H(q)\ddot q + C(\dot q, q) + G(q) = Bu + J^T \lambda,
    \label{eqn:manipulator_eqn}
\end{align}
where $H$, $C$, and $G$ are the mass matrix, Coriolis/centrifugal, and gravity terms respectively, and $B$ and $J$ map control inputs, $u$, and external forces $\lambda$ to generalized forces. If $B$ is full rank, we can compute the control input, $u$, that corresponds to an arbitrary acceleration, $\ddot q$. In the absence of external forces the inverse dynamics are given by
\begin{align}
    u = \text{ID}(\ddot q_\des) \triangleq B^{-1}\left [H(q)\ddot q_\des + C(\dot q, q) + G(q)\right],
\label{eqn:inverse_dynamics}
\end{align}
where $\ddot q_{\text{des}}$ is the desired acceleration. It is common to combine inverse dynamics and PD control with the control law
\begin{align}
u = \text{ID}\left (K_p (q_\des - q) + K_d(\dot q_\des - \dot q) \right).    
\label{eqn:inverse_dynamics_control}
\end{align}
In our experiments we set the damping gains to be the critical damping gains
\begin{align}
    K_d = 2\sqrt{K_p}.
\end{align}
Note that because the accelerations are mapped through the mass matrix, this controller doesn't suffer from the same scaling issues that arise in torque control and PD control.

\subsection{Impedance Control}
While the three previous controllers are all configuration space controllers, the impedance controller is a task space controller. Impedance control regulates the end effector dynamics to mimic a mechanical spring-damper system
,\begin{align}
    \ddot x + B (\dot x - \dot x_\des) + K (x - x_\des) = 0,
\end{align}
where $x$ is the end effector pose, $B$ is a damping matrix, and $K$ is a stiffness matrix. Taking two time derivatives of the end effector pose with respect to the joint coordinates, we get the relation $\ddot x = J\ddot q + \dot J \dot q$, where $J$ is the end effector Jacobian. 
Again, using the inverse dynamics, we can write the control law as
\begin{align}
    u &= \text{ID}\left( J^+ \left( (K(x_\text{des} - x(q)) + \dot x_\des - BJ\dot q - \dot J \dot q\right)\right),
\end{align}
where $J^+ = J^T(JJ^T)^{-1}$ is the pseudoinverse of the end effector Jacobian. The action space consists of $x_\des$ and $\dot x_\des$. In practice, we use $J^+ = J^T(JJ^T + \alpha I)^{-1}$, where $\alpha=1\times 10^{-6}$, to avoid large torques near kinematic singularities.

In our experiments, we use a simplified control law
\begin{align}
    u &= \text{ID}\left( J^+ \left( (K(x_\text{des} - x(q)) + \dot x_\des - BJ\dot q\right)\right),
\end{align}
because computing $\dot J \dot q$ in MuJoCo is computationally expensive. We can justify this as impedance control with an additional nonlinear damping term, $-J^+ \dot J \dot q$. Finally, since our arm has seven degrees of freedom, we damp out motions in the null space of the Jacobian with an additional damping term, $(I - J^+J)B_\text{null}\dot q$.
\section{Experiments}

\begin{figure}[thpb]
    \centering
    \includegraphics[width=0.5\textwidth]{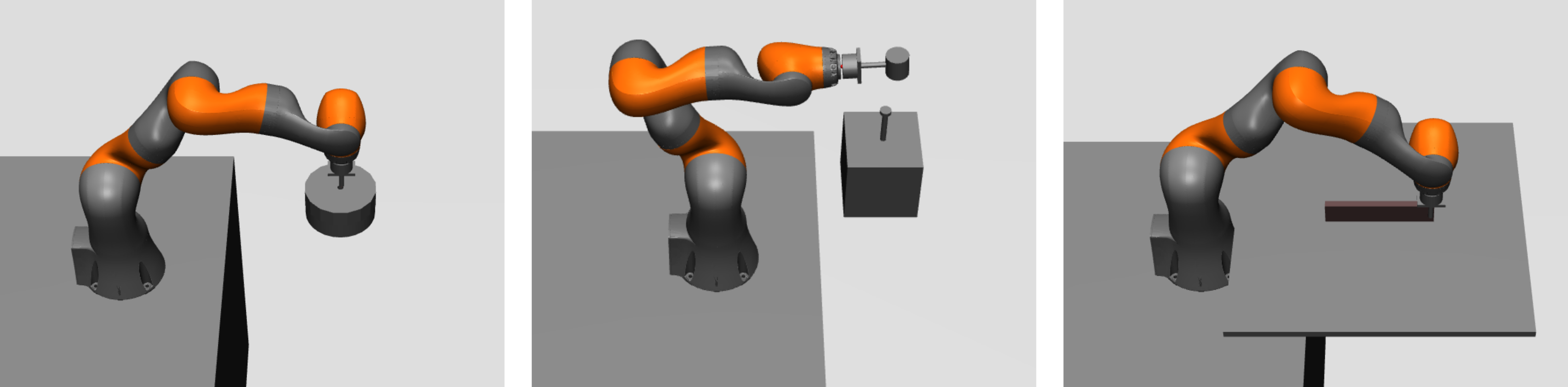}
    \caption{Simulated experiments from left to right: peg insertion, hammering, and a delicate pushing task. These experiments were selected to cover a range of manipulation skills including precision insertion, controlled impulses, and delicate interactions with other objects.}
    \label{fig:experiment_setup}
\end{figure}

We evaluated the performance of these action spaces on three representative manipulation tasks from the RL literature: peg insertion, nail hammering, and object pushing. Our experiments are performed on a model of a 7 degree of freedom Kuka IIWA 14 arm. Figure~\ref{fig:experiment_setup} illustrates the setup for each of the three experiments. All experiments were simulated in MuJoCo \cite{Todorov12}. The hyperparameters for each experiment can be found in the Appendix, and all of the code used for these experiments is publicly available in our GitHub repository \cite{Varin19}.
 
\subsection{Peg Insertion Environment}
Peg insertion is one of the most common manipulation tasks in robot assembly and requires navigating narrow bottlenecks in configuration space. This experiment consists of inserting a peg, rigidly affixed to the end effector, into a hole with 2\,mm clearance. In order to provide a generous comparison with the torque controller, gravity is disabled in the simulation. The observations consist of joint positions and velocities, the relative pose between the tip of the peg and the bottom of the hole, and the velocity of the end effector. The reward function is quadratic in the peg tip distance, $\Delta p$, and the orientation error, $\Delta \theta$,
\begin{align}
    r = -||\Delta p||^2 - ||\Delta \theta||^2.
\end{align}
The peg tip distance is computed from the relative position between the hole and the peg tip and the orientation error is the relative angle between the peg orientation and the hole orientation.

\subsection{Hammering Environment}
Hammering requires the controlled accumulation and delivery of momentum to a specific location in task space. The ``nail'' is 20\,cm long and experiences static friction that can resist 20\,N of force; we found this to be sufficiently high to require multi-impact hammering strategies rather than brute force pushing strategies. Similarly to the insertion task, gravity was disabled in order to learn useful strategies with the torque controller. The observations consist of joint positions and velocities, the nail position and velocity, the relative pose between the hammer face and the head of the nail, and the end effector velocity. The reward function is linear in the nail height as well as the peg velocity,
\begin{align}
    r = -h_\text{nail} - \dot h_\text{nail}.
\end{align}
Note that this reward structure is sparse, meaning that it is possible for the robot to experience many episodes without receiving any rewards.

\subsection{Object Pushing Environment}
In this experiment, we consider the problem of pushing an object from one position to another without tipping it over. The object is a rectangular prism, 30cm long, 2cm in width, and 6cm in height, and it is initialized to be standing on its narrow side. The robot is equipped with the same peg-like end effector from the insertion task. Unlike the hammering and insertion experiments, gravity is required to keep the block in contact with the table and provide frictional forces, so rather than disabling gravity we add a gravity compensation torque to the torque controller as well as the PD controller. The observations consist of the joint positions and velocities, the relative pose between the end effector and the center of the block, and the end effector velocity. The reward function is
\begin{align}
    r = -||\Delta p_\text{block}|| - ||\Delta \theta_\text{block}|| - k_\text{h}||\Delta h_\text{peg}||^2 - k_\theta||\Delta \theta_\text{peg}||^2,
\end{align}
where $\Delta p_\text{block}$ and $\Delta \theta_\text{block}$ are the relative position and orientation of the block respectively. The desired pose, $\Delta h_\text{peg}$, is the difference in height between the height of the block and the height of the peg, and $\Delta \theta_\text{peg}$ is the angle between the axis of the peg and vertical. The last two terms encourage the end effector to remain in-plane with the block, and the coefficients $k_\text{h}$ and $k_\theta$ are small to minimize unnecessary bias.
\section{Results}
The results of each of our experiments are collected in Table \ref{table:results}. The success criteria for the three tasks are: 80\% average nail depth for the hammer task, 80\% success rate for the insertion task, and 80\% of the normalized distance to the goal for the pushing task. In all of our experiments, we found the impedance controller action space to have learned the fastest, often followed by the inverse dynamics controller, PD controller, and then the torque controller. The favorable performance of the impedance controller can be understood in terms of the simplifying effect it has on the underlying task dynamics.

\begin{table}[h!]
\centering
\bgroup
\def\arraystretch{1.2} 
\begin{tabular}{ |c|c|c|c|c|c| } 
 \hline
  & & Impedance & ID & PD & Torque\\
 \hline
 \multirow{3}{2em}{PPO} & Hammer    & \textbf{0.11} & 0.28 & 0.40 & 0.45 \\ 
                        & Insertion & \textbf{0.14} & 0.47 & 0.61 & 0.46 \\ 
                        & Pushing   & \textbf{0.21} & 0.30 & 0.23 & 0.59 \\ 
 \hline
 \multirow{3}{2em}{SAC} & Hammer    & \textbf{0.012} & 0.029 & 0.077 & 0.145 \\ 
                        & Insertion & \textbf{0.39} & 0.47 & 0.94 & 0.94 \\ 
                        & Pushing   & \textbf{0.18} & 0.24 & 0.90 & * \footnotemark{} \\ 
 \hline
\end{tabular}
\egroup
\caption{Steps taking before reaching success criteria (millions).}
\label{table:results}
\end{table}

\begin{figure*}[t]
    \centering
    \includegraphics[width=0.32\textwidth]{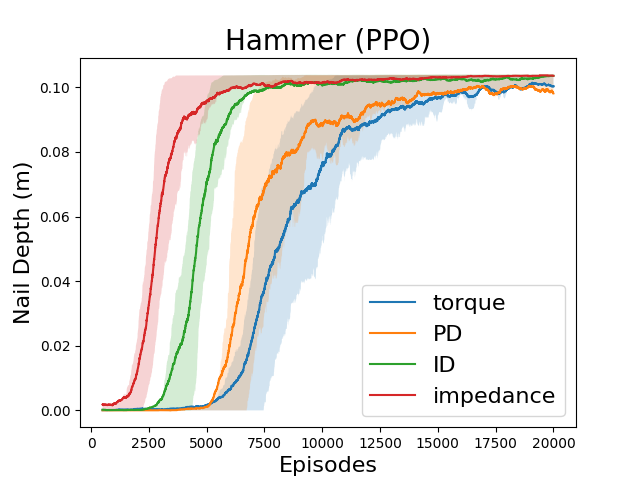}
    \includegraphics[width=0.32\textwidth]{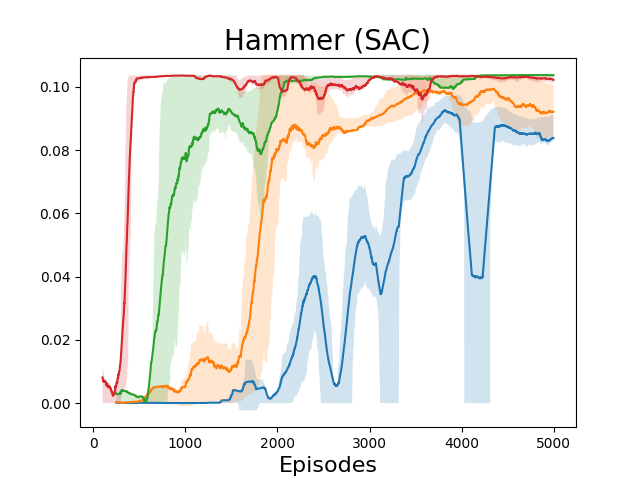}
    \includegraphics[width=0.32\textwidth]{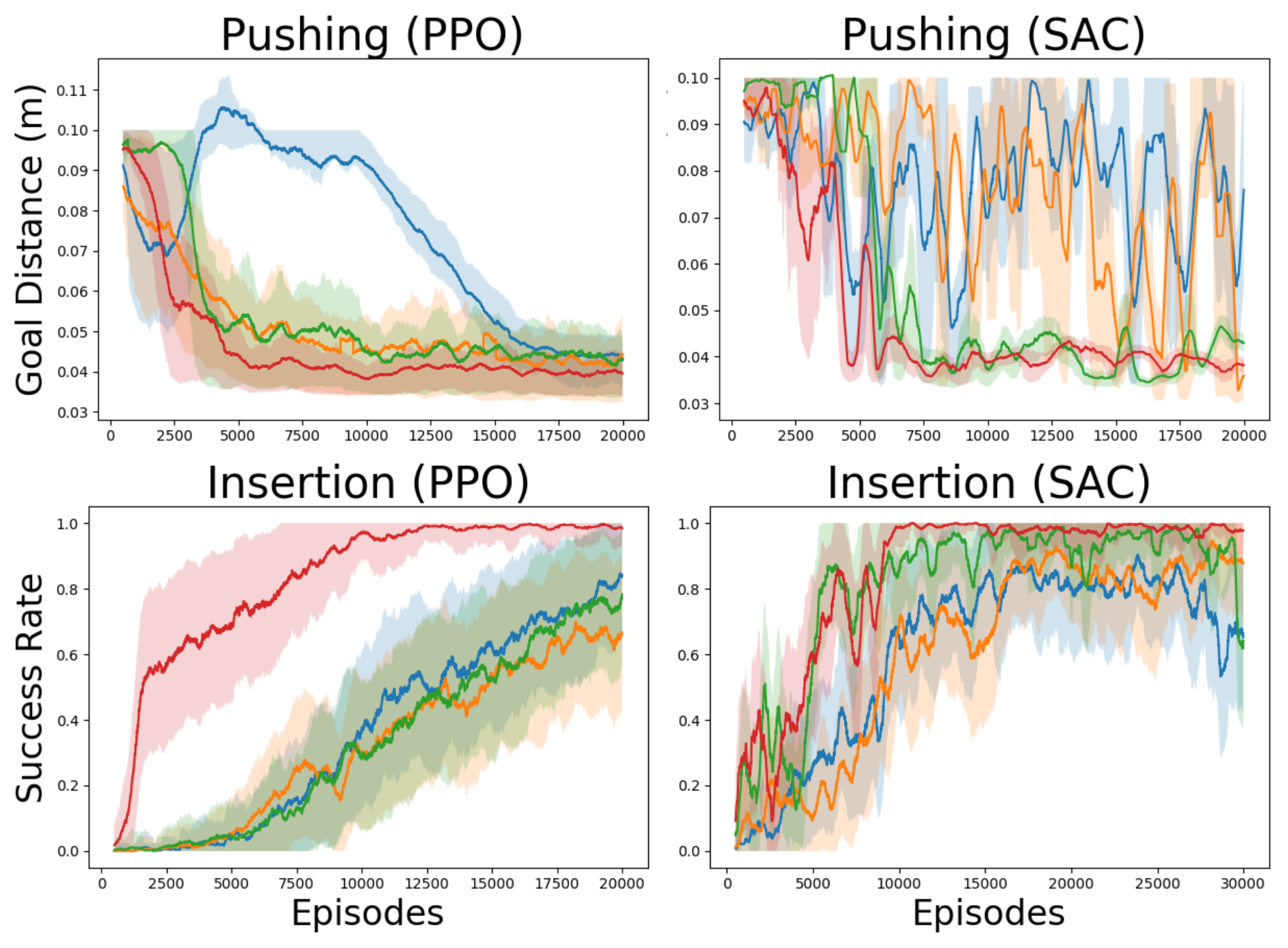}
    \caption{Training examples from learning hammering, pushing, and peg insertion using PPO and SAC.  Hammering with PPO (left) and SAC (middle) show how impedance control learns quicker than the other three controllers.  Similar trends are shown with PPO and SAC for the other two tasks (right).}
    \label{fig:learning_curves}
\end{figure*}

The role of compliance is highlighted in the pushing task. In order to keep the block upright while pushing it to the desired location, the interactions need to be gentle. Large contact forces can cause the block to topple, incurring a large cost for the remainder of an episode. Without the proper compliance, an algorithm can learn to avoid the block altogether. The impedance controller, the inverse dynamics controller, and the PD controller all have some inherent compliance and learn relatively quickly how to manipulate the block without knocking it over. The torque controller, however, has no inherent compliance and struggles with the task, learning three times slower than the impedance controller when trained with PPO and failing altogether when trained with SAC. There are also some qualitative differences between the learning curves for PPO and SAC. This can be attributed to the peculiarities of the two algorithms; PPO is gradient based and has asymptotic convergence guarantees, whereas Q-learning with function approximation, the basis for SAC, is known to suffer from instabilities during learning.

\footnotetext{The torque action space did not cross the 80\% threshold before 5 million steps.}

Somewhat surprisingly, the inverse dynamics controller shows improved performance over PD controller across five of the six experiments. It would have been reasonable to expect that, since both controllers are using a PD loop with comparable stiffnesses for feedback control, their performance would have been roughly equivalent on all tasks. It is noteworthy that this discrepancy is largest for the hammering task, where velocities are large, highlighting the sensitivity of these learning methods to the dynamics of the underlying system. During a hammer swing, inertial forces, rather than the feedback terms, can dominate the dynamics of the PD controller. The inverse dynamics controller, however, compensates for this inertial coupling between joints.

One of the biggest characteristics that sets the impedance controller apart, however, is that actions are specified in task space rather than configuration space. Because there is usually a straightforward mapping between the quantities in the reward function and the action space, the impedance controller often discovers useful behaviors much more quickly, without the extended exploration phase that the joint space controllers tend to exhibit. In the peg insertion task, for example, the space of successful joint configurations is quite complex, while the set of successful end effector poses is relatively simple. While the impedance control policy only needs to discover this simple set of poses, a joint space controller must also learn to solve an inverse kinematics problem to perform the insertion. Figure \ref{fig:learning_curves} shows this phenomenon most clearly in the case of the hammering task.  There, the impedance controller discovers the location of the nail very quickly and soon after completes the task. The joint space controllers require much more exploration before discovering the nail, then require longer training periods to refine the details of the task after the nail is located.

Qualitatively the policies learned in the different action spaces are also very different. The impedance, inverse dynamics, and PD controllers all exhibit slow, controlled motions whereas the torque controller tends to exhibit more aggressive motions with seemingly uncontrolled collisions with the environment. The impedance controller can exhibit odd artifacts near kinematic singularities during early training, but these are less noticeable in policies that have converged.

\begin{figure*}[t]
    \centering
    \includegraphics[width=0.9\textwidth]{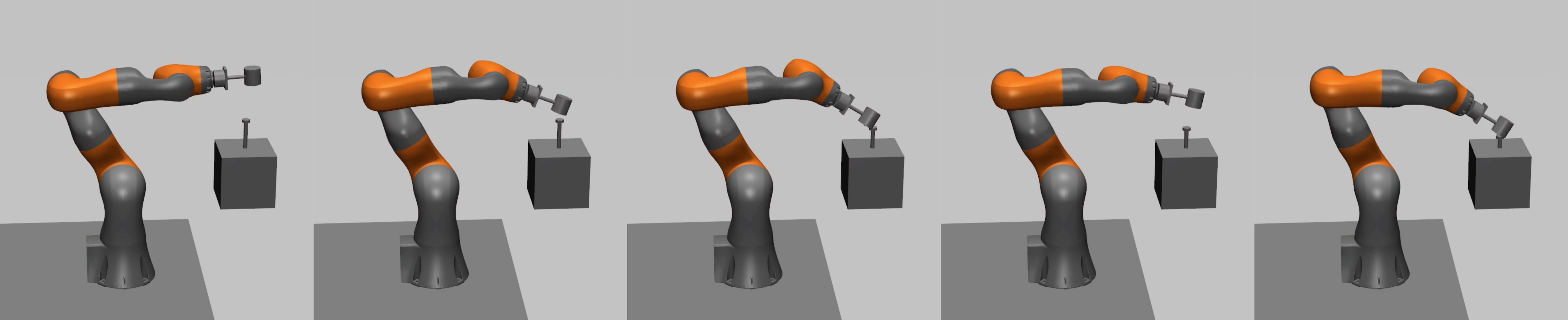}
    \caption{Frames from a learned hammer policy. The policy learns to take multiple swings at the nail in order to complete the task.}
    \label{fig:video_frames}
\end{figure*}

It is worth mentioning that we have gone to significant lengths to provide the best possible implementation of both the torque controller and the PD controller. Without care to scale the action space at each link by the mass of the child links, learning is significantly slower. In the presence of gravity, the torque controller learns none of the tasks, and in order for the PD controller to successfully complete the tasks, the gains would need to be increased by an order of magnitude, hindering its ability to be compliant in the presence of contact.


\section{Conclusions}
We evaluated learning performance for three dynamic manipulation tasks using four different action spaces: direct torque control, joint PD, inverse dynamics, and task-space impedance control. Action spaces defined in terms of torque control and PD control learned significantly slower than impedance control or inverse dynamics control on all of our experiments. These results are complementary to the results by Peng et al. in that PD control improves sample efficiency over direct torque control. However, we show that there can be additional benefit by wrapping the PD controller in an inverse dynamics routine, and that in many situations task-space impedance control has significant benefits over all of the joint space policies. 

Choosing the action space represents a trade-off between the engineering effort required to build and calibrate model-based controllers, and sample efficiency during learning. While it is sometimes desirable to prioritize simplicity of implementation, we expect that learning efficiency can usually be improved by incorporating model-based controllers that simplify some or all of the underlying task dynamics. 

One criticism of using model-based controllers in conjunction with model-free learning is additional dependence on having a model of the robot. While some of the details of the dynamics of manipulators---such as static friction in the joints---can be difficult to model, the inertial and kinematic model of the manipulators are often well known. The key point is that even if we cannot model all of the relevant dynamics in the learning task (e.g., we made no attempt to model the contact interaction between the robot and environment), there still may be value in exploiting what models are available to reduce the complexity of the learning problem, similar to previous findings on residual learning.

Finally, the experiments presented here do not address the potential for model mismatch between simulation and hardware. We expect policies learned with model errors to be able to compensate for the mismatch during learning, but it is unclear how this may affect learning performance and transfer of learned policies from simulation to hardware. We plan to investigate this further in future work.
\section*{Appendix}
\subsection{Learning Hyperparameters}
All experiments were carried out with consistent learning parameters for PPO and SAC.  PPO used a \texttt{gamma} of 1.0 and 2048 steps per gradient update.  All other parameters were set to the \texttt{stable\_baselines} default. The full hyperparameter list is given in Table \ref{table:hyperparameters}. Both PPO and SAC used two-layered feed-forward networks with 64 nodes per layer.

\begin{table}[h!]
\centering
\bgroup
\def\arraystretch{1.2} 
\begin{tabular}{ |l|l|c| }
  \hline
  & Parameter & Value \\
  \hline
   \multirow{8}{2em}{PPO} & MDP steps per update& 2048\\
   & learning rate & 0.00025\\
   & $\lambda$ for GAE & 0.95\\
   & value function cost & 0.5\\
   & entropy cost & 0.01\\
   & max gradient norm & 0.5\\
   & minibatches per update & 4\\
   & cliprange & 0.2\\
 \hline
   \multirow{7}{2em}{SAC} & learning rate & 0.0003\\
   & buffer size & 50000\\
   & entropy coefficient & adaptive\\
   & MDP steps between updates & 1\\
   & batch size & 64\\
   & Polyak update coefficient & 0.005\\
   & gradient steps per update & 1\\
 \hline
\end{tabular}
\egroup
\caption{Learning hyperparameters}
\label{table:hyperparameters}
\end{table}

The episode lengths were capped at 2 seconds for object pushing and peg insertion and 3 seconds for hamming. The policies were queried at 10Hz, and the low-level controllers operated at 100Hz across all experiments.







\label{appendix:hyperparameters}

\addtolength{\textheight}{-12cm}


\bibliographystyle{ieeetr}
\bibliography{bibliography.bib,extra.bib}

\end{document}